\documentclass{article}
\usepackage{codefuse_tech_report}
\usepackage[colorlinks = true,
            linkcolor = blue,
            urlcolor  = blue,
            citecolor = blue,
            anchorcolor = blue]{hyperref}   
\usepackage{microtype}
\usepackage{xurl}
\usepackage{booktabs}
\usepackage{enumitem}
\usepackage{multicol}
\usepackage{CJKutf8}
\usepackage{siunitx}
\usepackage{floatflt}
\usepackage{graphicx}
\usepackage{wrapfig}
\usepackage{authblk}
\usepackage{lipsum}
\usepackage{algorithm}
\usepackage{algorithmicx}
\usepackage{algpseudocode}
\usepackage{subfigure}
\usepackage{multirow}
\usepackage{pifont}  % 为了 \XSolidBrush
\usepackage{listings}

\algnewcommand{\LeftComment}[1]{\Statex \(\triangleright\) #1}

\usepackage{array}
\usepackage{amsmath}
\usepackage{amssymb}
\usepackage{mathtools}
\usepackage{amsthm}

\usepackage[capitalize,noabbrev]{cleveref}
\usepackage{adjustbox} % 引入adjustbox包

\theoremstyle{plain}

\theoremstyle{definition}

\theoremstyle{remark}

\colmfinalcopy

\usepackage[utf8]{inputenc}
\usepackage[T1]{fontenc}
\usepackage{caption} % Recommended for better caption control
\usepackage{adjustbox} % For vertical alignment
\usepackage{fontawesome5}

\usepackage{tcolorbox}
\tcbuselibrary{skins,breakable}

\tcbuselibrary{skins}

\usepackage{color}
\usepackage{xcolor}
\usepackage{soul} 

\definecolor{tred}{RGB}{251, 130, 132}
\definecolor{torange}{RGB}{247, 162, 116}
\definecolor{tyellow}{RGB}{251, 218, 140}
\definecolor{tgreen}{RGB}{127, 204, 181}
\definecolor{tblue}{RGB}{89, 177, 215}
\definecolor{insightblue}{RGB}{162, 210, 255}
\definecolor{questionred}{RGB}{255, 175, 204}

\title{GALLa: Graph Aligned Large Language Models for Improved Source Code Understanding}

\author{%
Ziyin Zhang\thanks{Equal Contribution.}$^{\phantom{*},1,2}$
~~Hang Yu$^{*,1}$
~~Shijie Li$^{1}$
~~Peng Di$^{1}$
\\

\vspace{-6pt}
\bf
~~Jianguo Li\thanks{Correspondence to: Jianguo Li \textless lijg.zero@antgroup.com\textgreater ~and Rui Wang \textless wangrui12@sjtu.edu.cn\textgreater.}$^{\phantom{\dagger},1}$
~~Rui Wang$^{\dagger, 2}$

\vspace{10pt}
$^1$Ant Group\ \ \ $^2$Shanghai Jiao Tong University\\
\vspace{10pt}
\hspace{-10pt}\faGithub ~\url{https://github.com/codefuse-ai/GALLa}\\
\hspace{-10pt}~~~~~~~~\includegraphics[width=1em,height=1em]{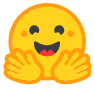} ~\url{https://huggingface.co/datasets/codefuse-ai/GALLa}\\
}

\colmfinalcopy % Uncomment for camera-ready version, but NOT for submission.

\begin{document}

\maketitle

% \begin{figure}[h!]
%     \centering
%     % \includegraphics[width=0.5\linewidth]{overview.pdf}
%     \caption{Put a figure here.}
%     \label{fig:overview}
% \end{figure}

\begin{abstract}
Programming languages possess rich semantic information - such as data flow - that is represented by graphs and not available from the surface form of source code. Recent code language models have scaled to billions of parameters, but model source code solely as text tokens while ignoring any other structural information. Conversely, models that do encode structural information of code make modifications to the Transformer architecture, limiting their scale and compatibility with pretrained LLMs. In this work, we take the best of both worlds with GALLa - Graph Aligned Large Language Models. GALLa utilizes graph neural networks and cross-modal alignment technologies to inject the structural information of code into LLMs as an auxiliary task during finetuning. This framework is both model-agnostic and task-agnostic, as it can be applied to any code LLM for any code downstream task, and requires the structural graph data only at training time from a corpus unrelated to the finetuning data, while incurring no cost at inference time over the baseline LLM. Experiments on five code tasks with seven different baseline LLMs ranging in size from 350M to 14B validate the effectiveness of GALLa, demonstrating consistent improvement over the baseline, even for powerful models such as LLaMA3 and Qwen2.5-Coder.
\end{abstract}

\section{Introduction}
In recent years, applying large language models (LLMs) to processing and generating source code has been a research topic of special interest in both natural language processing and software engineering community~\citep{2021HumanEval}. However, unlike natural languages, programming languages have rich semantics besides the lexical representation of source code, such as the path of execution, the flow of data, and the dependency of function calling. These semantics are represented by graph structures such as Abstract Syntax Tree (AST), Control Flow Graph (CFG), and Data Flow Graph (DFG). While enhancing LLMs with lexical representations of source code has the potential to boost their performance, the integration of these richer semantic constructs has yet to be fully realized~\citep{2023codesurvey}.

\begin{figure*}
    \centering
    \includegraphics[width=0.98\linewidth]{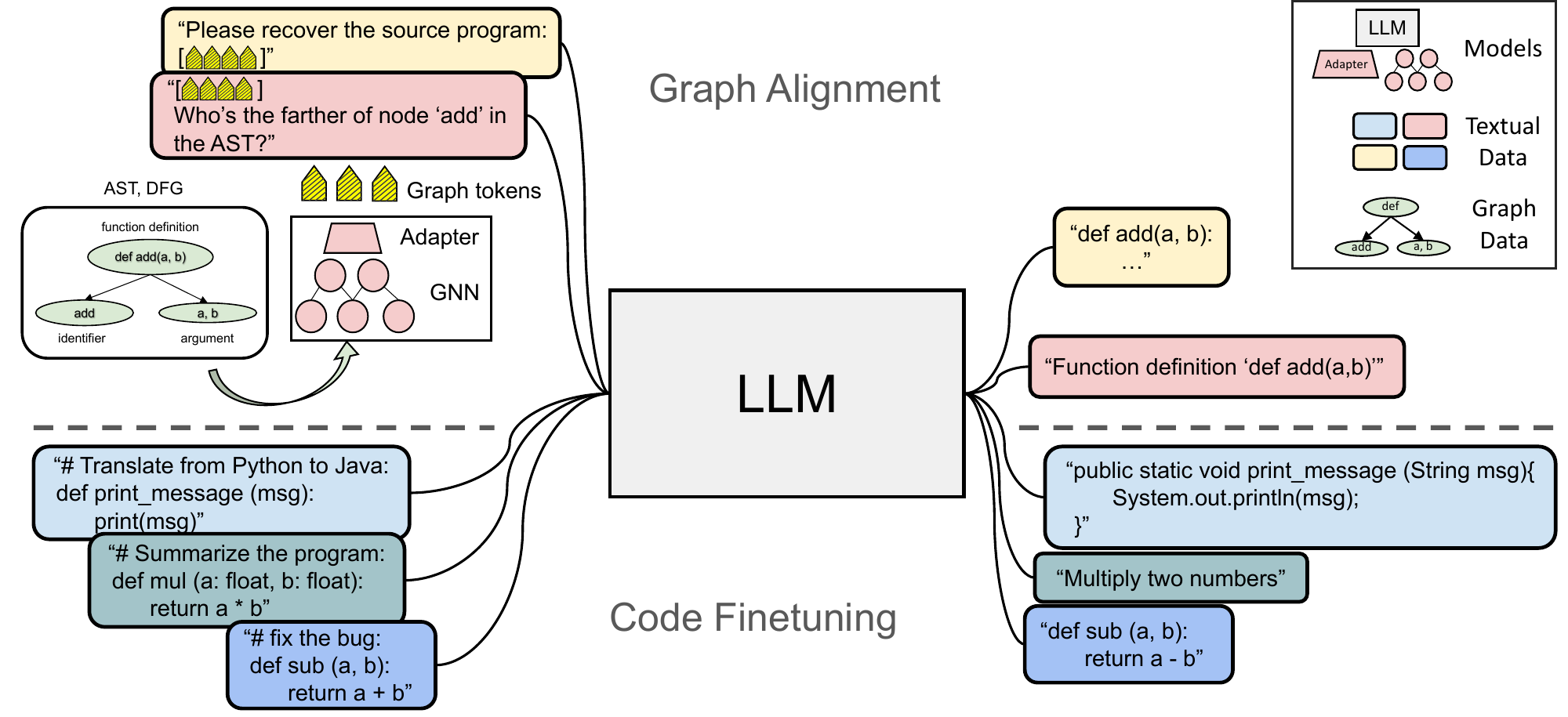}
    \caption{High-level overview of our method. The LLM is simultaneously trained on code downstream task data (such as code translation, code summarization, and code repair) and graph alignment data (graph-to-code generation, and question-answer pairs about the graph structures). The same-colored data on the left and right belong to the same input-output pair. The graph data is only required for the graph alignment tasks, but not the code finetuning tasks.}
    \label{fig:overview}
\end{figure*}

The major challenge of injecting code structures into code language models lies in the \textbf{incompatibility between structural graphs and large-scale pretrained language models}. On one end, following the scaling laws~\citep{2020scaling-law,2022Chinchilla}, some works have tried to improve the capability of language models in code processing by increasing their model size and pretraining data size, leading to code LLMs with tens or even hundreds of billions of parameters such as Code LLaMA~\citep{2023CodeLLaMA} and DeepSeek-Coder~\citep{2023DeepSeek-Coder,2024DeepSeek-Coder-V2}. However, these models are standard decoder-only Transformer language models~\citep{2017Transformer} trained with next token prediction, and are thus 
\textbf{unable to capture the semantics embedded in the structural graphs of code}.

On the other extreme, another line of research, represented by GraphCodeBERT~\citep{2020GraphCodeBERT} and TreeBERT~\citep{2021TreeBERT}, has focused on injecting graph structures into Transformer-based code language models. However, these methods either linearize the graphs into text tokens and are thus only applicable to simple tree-based graph structures~\citep{2022SPT-Code,2022UniXcoder}, or modify the Transformer architecture such as attention masks~\citep{2020GraphCodeBERT} and positional encodings~\citep{2021TPTrans} to encode graph information. \textbf{These modifications to the model structure make them incompatible with the large-scale pretrained decoder-only LLMs, and thus these experiments have been limited to a small scale}.

%Recently, researchers from the computer vision community have found a possible way to bridge 
Interestingly, insights from the computer vision community suggest a promising avenue for bridging the gap between different modalities: by utilizing a light-weight adapter to project the output features of a non-textual input processing model (such as an image encoder or a graph encoder) into the embedding space of language models, they are able to ground LLMs' understanding of text beyond the text modality, while in the meantime preserving LLMs' capability acquired during text-only pretraining~\citep{2023LLaVA,2023Qwen-VL,2023MiniGPT-4}.

Inspired by these works, we introduce GALLa: \textbf{G}raph \textbf{A}ligned \textbf{L}arge \textbf{La}nguage Models for code. We utilize a graph neural network (GNN) to process ASTs and DFGs extracted from source code, which are then projected into the language model's embedding space by a small adapter. The language model is trained to generate the source code conditioned on the graph information and to answer questions about the graph structure. These objectives align the language model's representation of code to the graph structures and impart it with a deeper understanding of the source code. As demonstrated in Figure~\ref{fig:overview}, our method is based on the transfer learning framework~\citep{2019T5} and separates graph alignment data from task-specific training data, thus preserving the general capability of the LLM acquired during pretraining and requiring no graph information about downstream task training or test data. 

Through extensive experiments on five code understanding and generation tasks, we validate the effectiveness of GALLa on seven distinct base LLMs ranging in size from 350M to 14B. GALLa brings consistent improvement over all baseline models, and even demonstrates abilities to generalize structural knowledge acquired during graph alignment to programming languages that are absent in the alignment data. All the data used in our experiments (both graph alignment and downstream tasks) are sourced from publicly available datasets\footnote{\url{https://huggingface.co/datasets/codefuse-ai/GALLa}}, and the complete code for reproducing our results is released in \url{https://github.com/codefuse-ai/GALLa}.
\section{Related Work}
Existing works that utilize structural graphs to enhance code language models can be categorized into three types: 1) modifying attention masks to encode graph information, 2) integrating graphs into the textual input, and 3) enhancing positional encodings with graph information.

The first category is represented by GraphCodeBERT~\citep{2020GraphCodeBERT}, which finetunes CodeBERT~\citep{2020CodeBERT} on concatenated source code and DFGs. Attention masks in the model are modified to reflect the graph structures: a node token $v_a$ is allowed to attend another node token $v_b$, only if there is an edge from $v_b$ to $v_a$ in the graph, while a node $v$ and a source code token $c$ can attend to each other only if they correspond to each other. StructCoder~\citep{2022StructCoder} also modifies attention masks similarly to encode the relations between source code tokens and AST, DFG tokens.

For the second category, TreeBERT~\citep{2021TreeBERT} is an encoder-decoder model where each input token to the encoder is a node's constituent path in the AST, while the decoder is trained to generate the source code. Many other works in this category, including SynCoBERT~\citep{2021SynCoBERT}, SPT-Code~\citep{2022SPT-Code}, and UniXcoder~\citep{2022UniXcoder}, map ASTs into text sequences (for example, by depth-first traversing) as model input. However, this method only applies to simple graph structures such as AST, but not more complex ones such as DFG, which may include loops.

In the third category, \citet{2021TPTrans} proposed TPTrans, which applies a recurrent sequence encoder to encode the relative paths between terminal nodes in ASTs and uses the results as relative positional encodings in self-attention modules, while the absolute paths from terminal nodes to the root node are similarly used as absolute positional encodings. \citet{2022TreeTransformer} used a list of 2-d coordinates and an embedding lookup table to describe the positions of AST nodes instead, and also used these embeddings to enhance both relative and absolute positional encodings in Transformer.

Despite their contributions, \textbf{all these approaches focus primarily on the encoder of the Transformer and are not fully compatible with decoder-only LLMs.} For instance, in the first group, the presence of cycles in the graphs means that the corresponding attention mask cannot retain a lower-triangular format, as required by LLMs. Adapting LLMs to these graph-structure-based attention masks would necessitate extensive retraining due to the mask's inconsistency with the original causal language modeling objectives. In contrast, the proposed GALLa method processes graph information externally, allowing the LLM architecture to remain unmodified.

Apart from these graph-enhanced models, there is another model TransCoder-IR~\citep{2022TransCoder-IR}, which utilizes LLVM intermediate representation (IR) to ground the model's understanding of code by generating source code from IR and vice versa. Like our method, TransCoder-IR only uses IR for alignment at training time but does not need it at test time. However, intermediate representations are also text tokens and provide limited structural information.
% represented in the same modality as the source code (i.e. text modality) and provide a different type of information compared with structural graphs.
\section{Method}\label{sec:method}
\begin{figure*}
    \centering
    \includegraphics[width=0.98\linewidth]{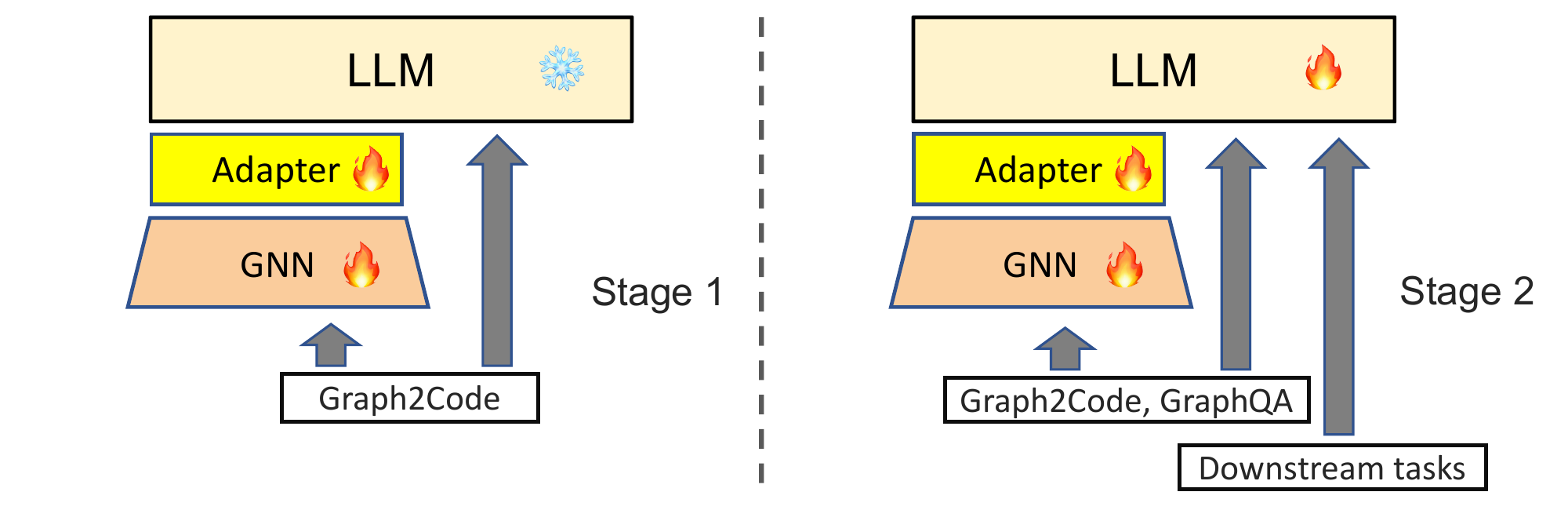}
    \caption{Illustration of the two-stage training schemes. In the first stage (left), the LLM's weights are frozen, and the GNN and adapter are pretrained on the Graph2Code task. In the second stage (right), the LLM is unfrozen and trained together with the GNN and adapter on the graph alignment tasks, while also simultaneously trained along (without the GNN and adapter) on downstream finetuning data. For Graph2Code and GraphQA tasks, the graph in a data sample goes into GNN, while the texts (including instruction and answer) in the sample go directly into the LLM.}
    \label{fig:training}
\end{figure*}

The objective of GALLa is to align language models to the implicit structures of code (represented by AST and DFG - see more details in Appendix~\ref{sec:appendix-graphs}) when finetuning on code-related tasks, and an overview of the method is provided in Figure~\ref{fig:overview}. In this section, we elaborate on two key challenges: \textbf{1) how to input graph information into LLMs}, and \textbf{2) how to design training tasks so that LLMs can learn about such graph information}. For the first challenge, we design a model that consists of three modules: GNN encoder, adapter, and LLM decoder~(Section~\ref{sec:method-model}). For the second challenge, we propose a two-stage training scheme: graph encoder pretraining and graph-LLM alignment~(Section~\ref{sec:method-training} and Figure~\ref{fig:training}). The notations used in this section are presented in Table~\ref{tab:notations}.

\begin{wraptable}{r}{0.5\linewidth}
    \centering
    \vspace{-0.2cm}
    \adjustbox{width=\linewidth,center}{
    \begin{tabular}{lm{6.5cm}}
    \toprule
        $\mathcal{G}$ & a graph \\
        $v$ & a node in the graph \\
        $e$ & an edge in the graph \\
        $n_v$ & number of nodes in the graph \\
        $n_e$ & number of edges in the graph \\
        $d_{\rm{node}}$ & node feature dimension of the graph data \\
        $d_{\rm{gnn}}$ & GNN hidden and output dimension \\
        $d_{\rm{lm}}$ & LLM hidden and embedding dimension \\
        $n_g$ & number of graph tokens in LLM's input \\
        $n_t$ & number of text tokens in LLM's input \\
    \midrule
        $V$ & node features (input to GNN, extracted by a text encoder) \\
        $E$ & edge indices (input to GNN) \\
        $H$ & contextual node features (output of GNN) \\
        $Q$ & query vectors (input to adapter) \\
        $X$ & embedding vectors (input to LLM) \\
        $Y$ & logits (output of LLM) \\
    \bottomrule
    \end{tabular}
    }
    \caption{Notations used in Section~\ref{sec:method}}
    \label{tab:notations}
\end{wraptable}

\subsection{Model Architecture}\label{sec:method-model}
\vspace{0.1cm}
\subsubsection{GNN Encoder}
To fully capture the rich information contained in the structural graphs such as loops, node degrees, and edge directions, we first process the graphs with a graph neural network (GNN) to extract node information. For a graph $\mathcal G$ with $n_v$ nodes and $n_e$ edges, a text encoder is used to extract node features from the corresponding source code of each node. These features are used to construct the node feature matrix $V\in\mathbb{R}^{n_v\times d_{ \rm{node}}}$, which are fed into the GNN along with the edge matrix $E\in\mathbb{Z}^{n_e\times2}$, where each element is a node index. The output of the GNN is the contextual node representations $H\in\mathbb{R}^{n_v\times d_{ \rm{gnn}}}$.

In this module, the text encoder can be any code embedding model such as CodeT5+~\citep{2023CodeT5+}. The GNN can be any convolution-based or self-attention-based directed GNN such as MagNet~\citep{2021MagNet} or DUPLEX~\citep{2024DUPLEX}, as DFG is a type of directed graph.

\subsubsection{Adapter}
The outputs of GNN are projected by an adapter into the LLM's embedding space. Following Qwen-VL~\citep{2023Qwen-VL}, we use a single layer of cross-attention as the adapter, where the GNN's outputs $H$ serve as keys and values, and the queries $Q\in\mathbb{R}^{n_g\times d_{\rm{lm}}}$ are $n_g$ learnable vectors:
\begin{equation}
    X_g =  \rm{CrossAttn}(q=Q, k=H, v=H).
\end{equation}

Alternative to the cross-attention layer, the adapter may also be a multi-layer perception (MLP), as used by LLaVA~\citep{2023LLaVA}, which applies projection independently to each node. The main difference between cross-attention and MLP is that cross-attention allows for information exchange between the nodes, while MLP is applied independently to each node. Thus, we choose cross-attention as the adapter in the main results, and experiment with MLP in the ablation studies in Section~\ref{sec:experiments-ablation}.

% Alternative to the cross-attention layer, the adapter may also be a multi-layer perception (MLP), as used by LLaVA~\citep{2023LLaVA}, which applies projection independently to each node. However, we observed inferior performance on MLP compared with cross-attention in preliminary experiments. We hypothesize that this is because cross-attention allows for information exchange between the nodes, thus allowing the module to select and extract features that are more relevant.

\subsubsection{LLM Decoder}
The adapter's outputs $X_g\in\mathbb{R}^{n_g\times d_{ \rm{lm}}}$ are $n_g$ embedding vectors, which we dub ``graph tokens''. Any other text in the LLM's input (as shown at the top of Figure~\ref{fig:overview}) is first tokenized and passed into the LLM's embedding layer to obtain $n_t$ text embeddings $X_t\in\mathbb{R}^{n_t\times d_{ \rm{ \rm{lm}}}}$, and then concatenated with the graph tokens to form the input to the LLM's Transformer layers:
\begin{equation}
    X = [X_g, X_t]\footnote{Depending on the actual input text, the graph tokens can also be placed after the text tokens or even inserted in the middle of text tokens, as illustrated in Figure~\ref{fig:overview}.}\in\mathbb{R}^{(n_g+n_t)\times d_{ \rm{lm}}}.
\end{equation}

The LLM's output logits $Y\in\mathbb{R}^{(n_g+n_t)\times d_{ \rm{lm}}}$ are then used to compute cross-entropy loss with next token prediction (i.e. causal language modeling). However, the loss is masked on the graph tokens and only computed on the text tokens.

\subsubsection{Model Choice} Lastly, we emphasize that GALLa is a framework for bridging the text modality and graph modality, and each of the three modules in GALLa can be instantiated with different models. The choice of GNN - e.g. directed or undirected - depends on the properties of graph data, while the choice of LLM depends on application scenarios - e.g. monolingual or multilingual, general-purposed or domain-specialized.

\subsection{Training Procedure}\label{sec:method-training}

Motivated by the pretraining + instruction finetuning paradigm in LLMs, we divide the training of GALLa into two stages. The first stage is self-supervised (continue) pretraining, where only AST/DFG data extracted from source code are used. The second stage is supervised finetuning, where GraphQA data are collected by designing graph-related questions and manually crafting question-answer templates.

\vspace{0.1cm}
\subsubsection{Stage 1: Graph Encoder Pretraining}
In GALLa, the LLM is initialized from a pretrained LLM checkpoint such as LLaMA~\citep{2024LLaMA3} or StarCoder~\citep{2023StarCoder}, while both the GNN and the adapter are randomly initialized. Thus, to prevent the newly initialized GNN and adapter from disrupting the LLM's pretrained representations, we fix the LLM's weights in stage 1, and update only the GNN and the adapter. In this stage, the model is trained with graph-to-code generation~(Graph2Code), where the model reconstructs a graph's corresponding source code tokens $X_t$ based on the graph tokens $X_g$ by maximizing the probability $P(X_t|X_g)$. Similar to visual instruction tuning~\citep{2023LLaVA}, this stage can be understood as training a ``graph tokenizer'' for the frozen LLM.

\vspace{0.1cm}
\subsubsection{Stage 2: Graph-LLM Alignment}
In the second stage, we aim to align the LLM's pretrained representations of source code to the structural graphs and deepen their understanding of code structures. In this stage, the LLM is unfrozen, and all three modules are updated together. The graph alignment tasks in this stage include Graph2Code (same as stage 1), and graph question answering (GraphQA). In GraphQA, the language model answers questions about a graph's structures, such as predicting whether there is an edge between two given nodes, or predicting the children of a given nodes. Formally, the model is trained to maximize the probabilities of the answer text tokens $X_a$ conditioned on the graph tokens $X_g$ and the question text tokens $X_q$: $P(X_a|X_g, X_q)$.

Since the ultimate goal of aligning LLMs to code graph structures is to enhance their performance on related downstream tasks, the model is simultaneously trained on the graph alignment data and downstream task data in this stage. However, we emphasize that the GNN and adapter are only used for the graph alignment tasks. No graph information is required of the downstream task data, as it goes directly into the LLM in the form of question-answer pairs, as shown at the bottom of Figure~\ref{fig:overview}.

\vspace{0.1cm}
\subsubsection{Inference}
After aligning the source code representation of LLMs to code graph structures in training stage 2, at inference time the graph encoder and adapter are discarded, so that the LLM can respond to user queries using its internal knowledge without any loss of speed compared with the base LLM. We choose not to encode the code in user queries with GNN at inference time because 1) it would require extracting AST and DFG from the code first, resulting in an unreasonable inference latency; 2) extracting AST and DFG requires that the input code is syntactically correct and complete, which is not true for many downstream tasks such as defect detection and code repair.

\vspace{0.1cm}

\section{Experiments}
\begin{table*}[t]
    \centering
    \adjustbox{width=\textwidth,center}{
    \begin{tabular}{ccccccccc}
    \toprule
         & Translation & Clone Detection & Defect Detection & Summarization & Repair \\
    \midrule
        Language & Py$\to$Java/Java$\to$Py & Java & C & Py/Java/JS & Java \\
        Train Samples & 44K/30K & 300K & 25K & 265K/170K/62K & 111K \\
        Test Samples & 499/164 & 10K & 3K & 3K/3K/3K & 12K \\
        Metric & pass@1 & F1 & Accuracy & BLEU & Exact Match \\
    \bottomrule
    \end{tabular}
    }
    \caption{Statistics of the downstream task datasets (Py: Python, JS: JavaScript).}
    \label{tab:task-stats}
\end{table*}

In this section, we first discuss the details of our experiments (\ref{sec:experiments-data}, \ref{sec:experiments-models}), and then provide the results in \ref{sec:experiments-results} and \ref{sec:experiments-ablation}.

\subsection{Datasets}\label{sec:experiments-data}
\vspace{0.1cm}
\subsubsection{Graph Alignment}
For graph alignment, we used 240K Python programs and 75K Java programs from CodeNet~\citep{2021CodeNet}. For each program, we extracted one AST and one DFG using Program Structure Interface~\footnote{\url{https://plugins.jetbrains.com/docs/intellij/psi.html}}, which resulted in 626K code-graph pairs after removing empty graphs and source code files that are longer than 4096 tokens. More details about these AST and DFG can be found in Appendix~\ref{sec:appendix-graphs}.

For the Graph2Code task, we removed all Python programs where no main function can be found, resulting in 150K Java  graph-code pairs and 81K Python graph-code pairs.

For the GraphQA task, we designed three types of questions:

1) Edge prediction: given the graph tokens and the source code of two nodes, the model is tasked to predict where there is an edge between them. This task is constructed only on DFG data.

2) Parent prediction: given the graph tokens and the source code of one node, the model is tasked to predict the node's parent in the graph. This task is constructed on both AST and DFG data.

3) Child prediction: given the graph tokens and the source code of one node, the model is tasked to predict the node's children in the graph. This task is constructed on both AST and DFG data.

We sampled about 75K graph-question-answer tuples for each (language, graph type, question type) combination, which resulted in a total of 770K GraphQA data. Further details of these GraphQA tasks and their prompts are given in Appendix~\ref{sec:appendix-prompts}.

\vspace{0.1cm}
\subsubsection{Downstream Tasks}
For downstream code tasks, we consider both discriminative and generative tasks, including code translation, clone detection, defect detection, code summarization, and code repair. For code translation, we use the CodefuseEval~\citep{2023CodeFuse} benchmark; for the other four tasks, we use the CodeXGLUE~\citep{2021CodexGLUE} benchmark. Among these tasks' data, we downsampled the original data of code summarization and clone detection due to computation constraints, and filtered out 14K Java$\to$Python samples in the training set of code translation where the target code does not follow Python coding conventions (e.g. no main function can be found). A summary of the final datasets is provided in Table~\ref{tab:task-stats}. For all tasks we use the original metric for evaluation.

In the main experiments, we consider the setting of multi-task finetuning (MFT), where the model is simultaneously trained on all five downstream tasks. However, we demonstrate with a small-scale experiment that GALLa can be also used for single-task finetuning (SFT).

\begin{table*}[t]
    \centering
    \adjustbox{width=\textwidth+0.2cm,center}{
    \begin{tabular}{llcccccc}
\toprule
Model & Setting & Trans (Ja2Py/Py2Ja) & Clone & Repair & Sum (Java/Py/JS) & Defect & Avg\\
\midrule
\multirow{3}{*}{CodeGen 350M} & Baseline & 40.2/42.3 & 94.4 & 9.0 & 13.6/8.8/12.5 & 56.9 & 34.7 \\
& G2C & 42.7/43.3 & 94.6 & 8.5 & 13.9/9.5/12.8$^*$ & 56.5 & 35.2 (+1\%) \\
& G2C + GraphQA & 50.0/45.3 & 93.9 & 8.6 & 14.0/9.5/12.7$^*$ & 58.7 & \textbf{36.6 (+5\%)} \\
\midrule
\multirow{3}{*}{StarCoder 1B} & Baseline & 0.6/0.0 & 40.3 & 0.4 & 7.8/3.2/5.1 & 54.2 & 14.0 \\
& G2C & 6.1/1.2 & 50.4 & 0.6 & 9.4/4.3/6.8 & 54.8 & 16.7 (+20\%) \\
& G2C + GraphQA & 6.1/1.2 & 65.8 & 0.3 & 9.5/4.5/7.4 & 56.5 & \textbf{18.9 (+36\%)} \\
\midrule
\multirow{3}{*}{Phi-1 1.3B} & Baseline & 73.8/53.3 & 93.5 & 10.1 & 14.6/11.3/13.8 & 61.1 & 41.4 \\
& G2C & 67.7/64.3 & 94.3 & 10.9 & 15.1/11.1/13.4$^*$ & 60.4 & 42.2 (+2\%) \\
& G2C + GraphQA & 72.0/66.3 & 94.9 & 10.2 & 15.2/11.7/13.6$^*$ & 60.3 & \textbf{43.0 (+4\%)} \\
\midrule
\multirow{3}{*}{Qwen2.5-Coder 1.5B} & Baseline & 35.4/73.8 & 95.2 & 13.2 & 14.4/10.8/13.6 & 58.4 & 39.3 \\
& G2C & 64.0/77.4 & 95.5 & 15.7 & 14.7$^*$/11.5/13.4 & 60.3 & 44.1 (+12\%) \\
& G2C + GraphQA & 65.2/76.0 & 95.7 & 15.0 & 15.0/11.3/14.3 & 61.3 & \textbf{44.2 (+12\%)}\\
\midrule
\multirow{3}{*}{LLaMA2 7B} & Baseline & 59.7/56.1 & 40.0 & 0.8 & 2.0/1.5/1.8 & 53.4 & 26.9 \\
& G2C & 69.5/59.1 & 41.5 & 1.0 & 2.4/1.7/1.9$^*$ & 53.5 & \textbf{28.8 (+7\%)} \\
& G2C + GraphQA & 64.0/53.3 & 39.0 & 1.6 & 2.2/1.6$^*$/2.0 & 55.2 & 27.4 (+2\%) \\
\midrule
\multirow{3}{*}{LLaMA3 8B} & Baseline & 80.5/74.8 & 94.7 & 12.1 & 14.2/11.2/12.3 & 56.7 & 44.6 \\
& G2C & 80.5/77.0 & 94.9 & 13.4 & 14.0$^*$/11.7/11.8 & 56.7 & 45.0 (+1\%) \\
& G2C + GraphQA & 80.5$^*$/78.2 & 94.9 & 13.6 & 14.5/11.7/12.9 & 57.1 & \textbf{45.4 (+2\%)} \\
\midrule
\multirow{3}{*}{Qwen2.5-Coder 14B} & Baseline & 38.4/82.1 & 94.3 & 12.7 & 18.0/20.2/14.5 & 58.5 & 42.3 \\
& G2C & 50.0/82.8 & 95.1 & 13.1 & 18.1$^*$/19.9/15.7 & 59.1 & 44.2 (+4\%) \\
& G2C + GraphQA & 64.0/83.4 & 94.4 & 13.8 & 17.9$^*$/20.5$^*$/15.9 & 58.0 & \textbf{46.0 (+9\%)}\\
\bottomrule
    \end{tabular}
    }
    \caption{Main results. For each model, the first row is baseline LLM finetuned on downstream task data only; the second row is GALLa finetuned on downstream task and Graph2Code data; the third row is GALLa finetuned on downstream tasks, Graph2Code, and GraphQA data. Relative performance increases w.r.t. baseline are given in parentheses in the last column. All differences from the baseline performance are statistically significant ($p<0.1$) except for those marked with $^*$ (the complete results are given in Appendix~\ref{sec:appendix-stat}).}
    \label{tab:main-mft}
\end{table*}

\begin{table*}[t]
    \centering
    \adjustbox{width=\textwidth,center}{
    \begin{tabular}{lccccccc}
\toprule
& Trans (Ja2Py/Py2Ja) & Clone & Repair & Sum (Java/Py/JS) & Defect & Avg$_{all}$ & Avg$_{java,py}$\\
\midrule
Baseline & 76.8/66.1 & 95.5 & 13.5 & 15.2/11.3/14.2 & 65.7 & 44.8 & 46.43\\
G2C & 79.3/68.9 & 95.8 & 15.1 & 15.2/11.7/14.1 & 64.8 & \textbf{45.6}$_{+2\%}$ & \textbf{47.67}$_{+3\%}$ \\
G2C+GQA & 75.6/69.5 & 95.7 & 13.6 & 15.3/11.7/13.9 & 59.9 & 44.4$_{-1\%}$ & 46.90$_{+1\%}$ \\
\bottomrule
    \end{tabular}
    }
    \caption{Results of single-task finetuning with Phi-1. Relative performance increases w.r.t. baseline are given in subscripts in the last two columns. The penultimate column is the average of all tasks, while the last column is the average of all Java and Python tasks. G2C: Graph2Code. GQA: GraphQA.}
    \label{tab:main-sft}
\end{table*}

\subsection{Models and Training}\label{sec:experiments-models}
\vspace{0.1cm}
\subsubsection{Model}
We use a DUPLEX~\citep{2024DUPLEX} with 1024 hidden states and 7M parameters as the GNN, and a cross-attention layer with randomly initialized learnable queries as the adapter. For the LLM, we consider six distinct models of varying sizes: CodeGen 350M~\citep{2022CodeGen}, StarCoder 1B~\citep{2023StarCoder}, Phi-1 1.3B~\citep{2023Phi-1}, LLaMA2 7B~\citep{2023LLaMA2} LLaMA3 8B~\citep{2024LLaMA3}, and Qwen2.5-Coder 1.5B \& 14B~\citep{2024Qwen2.5-Coder}. All of these models are prerained (partially) on source code data, and demonstrate strong performance on code-related downstream tasks.

\subsubsection{Training}
For the first stage training of GNN and adapter, we train the model on the graph data for 15 epochs with learning rate 1e-4, 1K warmup steps, weight decay 0.1, 240 global batch size, AdamW optimizer~\citep{2017AdamW}, and ZeRO stage 2~\citep{2019ZeRO}. The training takes place on two machines, each equipped with 8 A100 80G GPUs.

The second stage of training largely follows the same setting but with a smaller learning rate (5e-5) and a smaller global batch size (96). The model is trained for 5 epochs on the mixture of downstream task data and graph alignment data in stage 2, and the checkpoint with the lowest validation loss on the downstream tasks is chosen for evaluation. Among the different LLMs, CodeGen, StarCoder, Phi-1, and Qwen2.5-Coder 1.5B are trained in full scale, while LLaMA2, LLaMA3, and Qwen2.5-Coder 14B are trained using LoRA~\citep{2021LoRA} with rank 64. All training in this stage takes place on a single machine with 8 A100s.

As a baseline, the LLM is finetuned on only the downstream task data using the same hyper-parameters as stage 2 training.

\subsection{Results}\label{sec:experiments-results}
The results of applying Graph2Code and GraphQA alignment to the code finetuning process of five models are presented in Table~\ref{tab:main-mft}. Graph alignment brings consistent improvement over the baseline model, especially on the weaker backbones such as StarCoder, increasing the average performance on five tasks by up to 36\%.

% increasing the average performance on five tasks by up to 36\%. Comparing the results between the five models, we find that GALLa's effectiveness is most prominent on weaker backbone models such as StarCoder, while relatively less notable on the strongest model LLaMA3. The statistical significance tests of these results are given in Appendix~\ref{sec:appendix-stat}.

Notably, while our graph alignment data include only Python and Java programs, from Table~\ref{tab:main-mft} we observe that they can even improve six of the seven models' performance on tasks in other languages - code summarization in JavaScript, and defect detection in C. This showcases that the knowledge about code structures acquired in GALLa can be generalized across programming languages, as learning to align to Python and Java structural graphs improves the finetuning performance of downstream tasks in other languages.

In Table~\ref{tab:main-sft}, we also present the results of single-task finetuning with Phi-1. In this setting, we find that Graph2Code still improves the average performance on all tasks by 2\%, and by 3\% on Python and Java tasks. On the other hand, GraphQA brings limited improvement, and even a large drop on the C defect detection task. We hypothesize that this is because the diverse data in GraphQA serve a similar role to instruction tuning for LLMs~\citep{2021T0,2022Flan}, where the benefits of cross-task transfer only start to manifest when the number of tasks is large. We find that models trained with single-task finetuning are prone to hallucinations, where they do not follow instructions in test samples but output answer templates learned from the GraphQA data instead (see Appendix~\ref{sec:appendix-sft-example} for examples).

\begin{table*}[t]
    \centering
    \begin{tabular}{lcccccc}
\toprule
& Trans (Ja2Py/Py2Ja) & Clone & Repair & Sum (Java/Py/JS) & Defect & Avg\\
\midrule
Baseline & 76.8/66.1 & 95.5 & 13.5 & 15.2/11.3/14.2 & 65.7 & 44.8\\
G2C & 79.3/68.9 & 95.8 & 15.1 & 15.2/11.7/14.1 & 64.8 & \textbf{45.6} \\
G2C (AST only) & 76.8/66.9 & 95.8 & 13.6 & 15.1/11.9/13.8 & 65.2 & 44.9 \\
G2C (DFG only) & 79.3/68.3 & 95.8 & 13.7 & 15.1/11.9/13.8 & 66.3 & 45.5 \\
G2C (Code only) & 78.7/69.3 & 95.1 & 12.4 & 15.2/12.3/14.0 & 62.9 & 45.0 \\
\bottomrule
    \end{tabular}
    \caption{Ablation studies on the Graph2Code (G2C) task with Phi-1.}
    \label{tab:ablation}
\end{table*}

\begin{table*}[t]
    \centering
    \adjustbox{width=\textwidth,center}{
    \begin{tabular}{llcccccc}
\toprule
& Training & Trans (Ja2Py/Py2Ja) & Clone & Repair & Sum (Java/Py/JS) & Defect & Avg\\
\midrule
 & Baseline & 73.8/53.3 & 93.5 & 10.1 & 14.6/11.3/13.8 & 61.1 & 41.4 \\
\midrule
\multirow{2}{*}{MLP} & G2C & 73.2/66.9 & 95.2 & 11.7 & 15.1/11.6/14.1 & 61.6 & 43.7 \\
& G2C + GQA & 76.2/68.9 & 95.3 & 12.7 & 14.9/12.1/13.6 & 59.9 & \textbf{44.2} \\
\midrule
\multirow{2}{*}{MagNet} & G2C & 75.6/65.3 & 94.9 & 11.2 & 14.5/11.3/13.9 & 63.1 & 43.7 \\
& G2C + GQA & 74.4/65.1 & 95.0 & 12.8 & 15.0/12.4/14.1 & 63.2 & \textbf{44.0} \\
\bottomrule
    \end{tabular}
    }
    \caption{Experiments with Phi-1 using an MLP instead of a cross-attention layer as the adapter (top), and using MagNet instead of DUPLEX as the graph encoder (bottom).}
    \label{tab:mlp}
\end{table*}

\subsection{Ablation Studies}\label{sec:experiments-ablation}

% \begin{table*}[!t]
%     \centering
%     \begin{tabular}{lcccccc}
% \toprule
% & Trans (Ja2Py/Py2Ja) & Clone & Repair & Sum (Java/Py/JS) & Defect & Avg\\
% \midrule
% Baseline & 73.8/53.3 & 93.5 & 10.1 & 14.6/11.3/13.8 & 61.1 & 41.4 \\
% G2C & 75.6/65.3 & 94.9 & 11.2 & 14.5/11.3/13.9 & 63.1 & 43.7 \\
% G2C + GQA & 74.4/65.1 & 95.0 & 12.8 & 15.0/12.4/14.1 & 63.2 & \textbf{44.0} \\
% \bottomrule
%     \end{tabular}
%     \caption{Experiments with Phi-1 using MagNet instead of DUPLEX as the graph encoder.}
%     \label{tab:magnet}
% \end{table*}

\paragraph{Contribution of AST and DFG}
As an ablation study, we conducted experiments with Phi-1 and Graph2Code using either only the AST data or only the DFG data for graph alignment. The results are shown in Table~\ref{tab:ablation}. We find that DFG brings more improvement compared with AST, which we contribute to the fact that AST is more closely related to the surface form of source code and thus provides the model with less additional structural information, while DFG includes more complex structures - such as loops - that are more informative.

\paragraph{Contribution of the Graph Information}
Observing the previous results, one may raise the question: are these improvements indeed attributable to the alignment of code representations to graphs, or are they simply a result of additional computation expense during the finetuning of LLMs?

\begin{figure}[t]
    \centering
    \includegraphics[width=0.6\columnwidth]{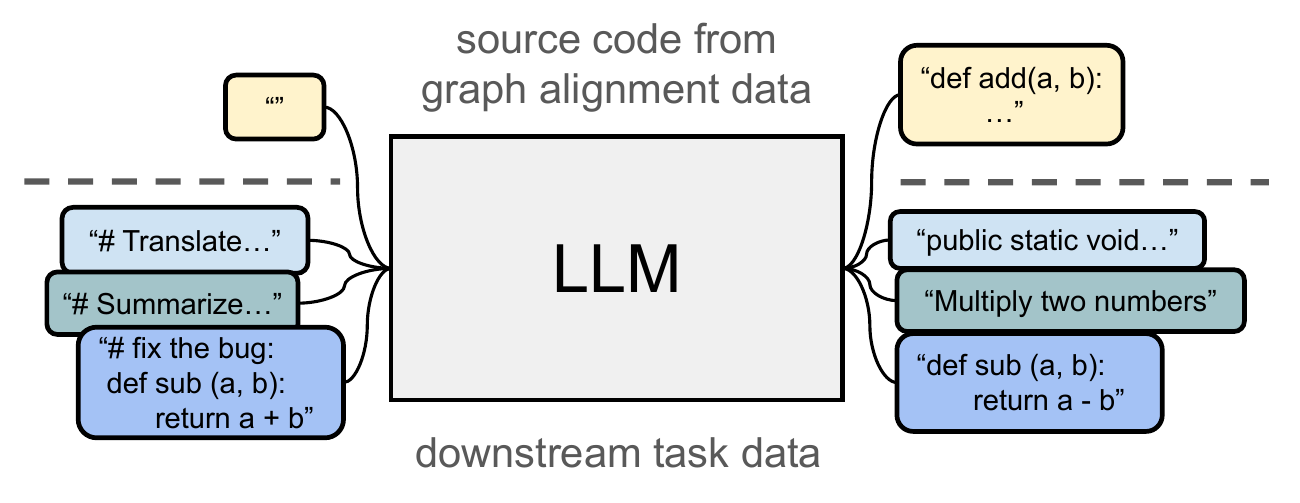}
    \caption{The setting of control experiment, where the model is trained on the same source code data as the Graph2Code task in Figure~\ref{fig:overview} but the graph information is not given, i.e. Graph2Code without the graph.}
    \label{fig:control}
\end{figure}

To answer this question, we conducted a control experiment where we additionally trained a model on the mixture of downstream tasks and the source code from the graph alignment data, but did not provide it with the graph information - in other words, Graph2Code with only the code but not the graph, as illustrated in Figure~\ref{fig:control}, which is similar in essence to continual pretraining on the source code data. The results are given in the last row in Table~\ref{tab:ablation}, which suggests that the graph information is indeed helping the LLM to better understand programs.

\paragraph{Different Adapters and Graph Encoders}
To show that the proposed GALLa framework can be applied to various GNN encoders and adapter modules, we also conducted another two sets of experiments in the MFT setting with Phi-1: based on the settings in the main experiments, we 1) replaced the cross-attention adapter with a 3-layer MLP (which has a similar parameter count to the cross-attention layer), and 2) replaced the graph encoder with MagNet~\citep{2021MagNet}. For these two experiments, we repeated the complete two stages of GALLa training, and the results are presented in Table~\ref{tab:mlp}. 

Compared with the results using cross-attention from Table~\ref{tab:main-mft}, we surprisingly find that the MLP adapter leads to a slightly better performance, which is counter-intuitive as MLP does not allow for information exchange between the nodes. We hypothesize that this is because with the same parameter count, MLP has more layers than cross-attention (3 vs. 1 in our case), thus having a stronger expressivity.
% The results of replacing DUPLEX with MagNet are given in Appendix~\ref{sec:appendix-magnet}.

Similarly, when using MagNet (which is convolution-based) instead of DUPLEX (which is self-attention-based) for finetuning Phi-1, the model's performance on downstream tasks is also slightly improved. These results of using different adapters and graph encoders, together with the experiments of using different LLMs~(Table~\ref{tab:main-mft}), confirm that GALLa is indeed a flexible framework where each of the three modules - GNN encoder, adapter, and LLM decoder - can be replaced as more advanced models are proposed in the future, leading to better performance on downstream tasks.

\section{Conclusion}
In this work, we present the conceptual designs, implementation details, and experimental results of GALLa - Graph Aligned Large Language Models for improved source code understanding. Unlike previous works that modify code language models' internal structures to enhance them with graph information, GALLa follows the cross-modality alignment paradigm and leaves the structure of the language models intact, making it applicable to any off-the-shelf code LLMs. By integrating GALLa as an auxiliary task on a separate graph alignment dataset in the finetuning stage of code LLMs, we require no graph information of the task-specific finetuning or evaluation data, thus incurring no additional computation cost compared with the baseline LLM at inference time. Our experiments validate the effectiveness of GALLa on various code downstream tasks with base LLMs of varying sizes, paving road for a new paradigm in integrating code structural graphs into language models and providing insights for future research in developing structure-aware code LLMs.

\bibliographystyle{colm2024_conference}
\bibliography{custom}

\appendix
\clearpage
\section{Introduction to Structural Graphs}\label{sec:appendix-graphs}
In this work, we refer to Abstract Syntax Trees (ASTs) and Data Flow Graphs (DFGs) as program structural graphs. These are graph data structures (i.e. nodes and edges) that specify the internal logic of programs. Each node in these graphs is a snippet of code within the entire program.

\begin{wraptable}{r}{0.5\linewidth}
    \centering
    \vspace{-0.2cm}
    \adjustbox{width=\linewidth,center}{
    \begin{tabular}{llll}
    \toprule
        idx & content & type & parent idx \\
    \midrule
        1 & \texttt{a, b} & \emph{arguments} & 0 (root) \\
        2 & \texttt{return a+b} & \emph{return statement} & 0 (root) \\
        3 & \texttt{a} & \emph{arguments} & 1 \\
        4 & \texttt{b} & \emph{arguments} & 1 \\
        5 & \texttt{a+b} & \emph{binary expression} & 2 \\
        6 & \texttt{a} & \emph{variable} & 5 \\
        6 & \texttt{b} & \emph{variable} & 5 \\
    \bottomrule
    \end{tabular}
    }
    \caption{An illustrative example AST for the toy program in Appendix~\ref{sec:appendix-AST}. Each row is a node in the graph.}
    \label{tab:example-ast}
\end{wraptable}

\subsection{AST}\label{sec:appendix-AST}
An AST is a tree representation of the syntactic structures of a program, with the root node being the entire program, and each leaf node being a single semantic unit in the program - such as an identifier (e.g. a variable, a class, a function) or an operator. Each non-leaf node in the tree is a combination of its children, indicating the structures of the code. As the name suggests, AST is an abstract representation of source code, in the sense that certain details such as white spaces, parentheses, and delimiters are omitted. Every node in the tree has an associated type attribute, such as \emph{assignment expression} or \emph{for loop}. A toy example in Python is provided below:
\begin{verbatim}
def add(a, b):
    return a+b
\end{verbatim}
In the AST for this example, the root node is a \emph{function expression} including the whole code snippet. The rest of the nodes are given in Table~\ref{tab:example-ast}.

\begin{wraptable}{r}{0.5\linewidth}
    \centering
    \vspace{-1cm}
    \begin{tabular}{lll}
    \toprule
        idx & from & to \\
    \midrule
        1 & \emph{variable} \texttt{a} & \emph{binary expression} \texttt{a+b}\\
        2 & \emph{variable} \texttt{b} & \emph{binary expression} \texttt{a+b}\\
        3 & \emph{arguments} \texttt{a} & \emph{variable} \texttt{a}\\
        4 & \emph{arguments} \texttt{b} & \emph{variable} \texttt{b}\\
    \bottomrule
    \end{tabular}
    \caption{An illustrative example DFG for the toy program in Appendix~\ref{sec:appendix-AST}. Each row is an edge in the graph.}
    \label{tab:example-dfg}
    \vspace{-1cm}
\end{wraptable}

\subsection{DFG}\label{sec:appendix-DFG}
A DFG is a graph representation of the variable use-define chains within a program at execution time. For a given program, its DFG shares the same set of nodes with its AST, while the edges indicate the data flow between variable-involved nodes. An example DFG of the previously shown program is provided in Table~\ref{tab:example-dfg}.

\subsection{Universal AST}
Typically, ASTs are specific to the programming language in question. In other words, different programming languages would have different types of nodes in their ASTs - for example, a \emph{for loop} node in Python may have different semantics from a \emph{for loop} in C++. However, in the context of our work, such language-specific specifications may be detrimental to the cross-language alignment (e.g. the code translation task) or generalization capabilities (i.e. generalizing to languages not present in the training data) of the models.

Thus, in this work we used a special type of AST (and DFG, as they share the same set of nodes): Universal AST (UAST). In UAST, the specifications of all node types are designed to be as language-independent as possible. Taking the nodes in Table~\ref{tab:example-ast} as an example, UAST abstracts language-specific node types into the most basic concepts shared by most programming languages, such as variable, binary expression, and return statement. In total, there are 43 node types in our graph data.

% \emph{for loops} from different languages are not distinguished from each other, and only the most basic semantics of a for loop is retained in this abstract node type.

% To improve the generalization capability of our model and to align the representations of different programming languages for certain cross-lingual tasks such as code translation, we used a special type of AST and DFG - universal AST (UAST) and universal DFG (UDFG), where different languages share the same set of node types to the possible extent (see more details in Appendix~\ref{sec:appendix-graphs}). In total, there are 43 node types in our graph data.

\clearpage
\section{GraphQA Prompts}\label{sec:appendix-prompts}
In our main experiments, we used three types of questions for GraphQA: edge prediction, parent prediction, and child prediction. For each type of question, we wrote about ten question templates and ten answer templates, as shown in Figure~\ref{fig:prompt-edge-prediction} to \ref{fig:prompt-child-prediction}. When constructing the data, one question template and one answer template are randomly chosen for each sample, as diverse question templates have been found to improve cross-task generalization capability in instruction finetuning~\citep{2021T0}. The placeholders in the templates are replaced with the actual node contents and types, and the instantiated prompt is then randomly placed before or after the graph tokens.

In our preliminary experiments, we found edge prediction data constructed from AST to bring no improvement, as connected nodes in the AST often have high textual overlap and can be trivially predicted. Thus we only use DFG data for the edge prediction task, while both AST and DFT data are used for the other two tasks. In early trials, we also experimented with and eventually discarded several other types of questions, including node classification (which can be easily done by only looking at the node's source code) and some tasks that involve counting, such as counting the number of nodes in the graph, the number of edges in the graph, and the number of node types in the graph, which prove to be too difficult for the model.

\lstset{
  basicstyle=\normalsize,
  columns=fullflexible,
  frame=single,
  breaklines=true,
  breakindent=0pt,
  moredelim=[is][\textbf]{\#\#\#\#\#\#}{\#\#\#\#\#\#},
}

\begin{figure*}[h]
\centering
\begin{lstlisting}
Question templates:
- In the graph, is there an edge from {node_type1} {node1} to {node_type2} {node2}?
- In the graph, is there an edge pointing from {node_type1} {node1} to {node_type2} {node2}?
- Please tell me if there is an edge pointing from {node_type1} {node1} to {node_type2} {node2} in this graph.
- Is there an edge from {node_type1} {node1} to {node_type2} {node2} in this graph?
- Does a connection exist from {node_type1} {node1} to {node_type2} {node2} in the graph?
- In this graph, do we have an edge leading from {node_type1} {node1} to {node_type2} {node2}?
- Is it true that {node_type1} {node1} is a predecessor of {node_type2} {node2} in this graph?

Answer templates 1 (positive):
- Yes, that is the case.
- Yes, there is an edge from {node_type1} {node1} to {node_type2} {node2}.
- Yes, there is an edge from {node_type1} {node1} to {node_type2} {node2} in this graph.
- Yes, there is an edge pointing from {node_type1} {node1} to {node_type2} {node2} in this graph.
- Affirmative, there exists an edge from {node_type1} {node1} to {node_type2} {node2}.
- Yes, that is the case. {node1} is directly connected to {node2}.

Answer tempaltes 2 (negative):
- No, that is not the case.
- No, {node_type1} {node1} is not linked to {node_type2} {node2} by any edge in this graph.
- No, there is no edge from {node_type1} {node1} to {node_type2} {node2}.
- No, such an edge is absent from the graph.
- The graph does not show {node_type1} {node1} as a predecessor to {node_type2} {node2}.
\end{lstlisting}
\caption{Prompts for the edge prediction GraphQA task.}
\label{fig:prompt-edge-prediction}
\end{figure*}

\begin{figure*}[h]
\centering
\begin{lstlisting}
Question templates:
- In the graph, what is the parent node of this {node_type}: {node}.
- What is the parent of {node_type} {node} in this graph?
- What is the parent node of {node_type} {node} in the graph?
- Based on the graph, identify the parent of {node_type} {node}.
- Based on this graph, identify the parent of this {node_type}: {node}.
- Identify the parent of {node_type} {node} in the graph.
- In the graph presented, what is the predecessor of {node_type} {node}?
- What node acts as the parent to {node_type} {node} in the graph displayed?
- Can you determine the parent node of {node_type} {node} in this graph?
- Which node is directly above {node_type} {node} in the hierarchy of the provided graph?
- What is the immediate ancestor of the {node_type} {node} in this graph?
- Regarding the graph, can you point out the parent of {node_type} {node}?
- In terms of graph theory, what is the parent of the {node_type} {node}?
- Who has the parental role for {node_type} {node} in the graph's topology?
- For {node_type} {node} in the given graph, which node supplies the incoming edge?

Answer templates 1 (has parent):
- In the given graph, the parent of the given {node_type} is {parent}, which is a {parent_type}.
- This {node_type}'s parent is the {parent_type} {parent}.
- The given {node_type}'s parent in the graph is the {parent_type} {parent}.
- The parent of {node_type} {node} in this graph is identified as {parent}, categorized as a {parent_type}.
- Node {parent}, a {parent_type}, serves as the parent to {node_type} {node} in the graph.
- As per the hierarchy, the {parent_type} node {parent} is the direct predecessor to {node_type} {node}.
- Upon inspection, it is clear that the parent of {node_type} {node} is the {parent_type} {parent}.
- The {node_type} {node} is immediately descended from {parent}, a {parent_type} in the graph.
- Within the nodal arrangement, {parent} is the progenitor to {node_type} {node}, having the classification of a {parent_type}.
- Tracing the edges leads to confirming {parent}, a {parent_type}, as the parent of {node_type} {node}."

Answer tempaltes 2 (no parent):
- This {node_type} has no parent in the graph.
- There is no edge pointing to this {node_type} in the given graph. Therefore it does not have any parent.
- Within this graph, {node_type} {node} does not have a parent node.
- {node_type} {node} stands without a parent in the graph's existing structure.
- No parent node is associated with {node_type} {node} in the provided graph.
- A review of the graph establishes that there is no preceding node to {node_type} {node}; it has no parent.
- In this graph topology, {node_type} {node} is an orphan node with no parent.
- There is no edge incoming to {node_type} {node}, indicating the absence of a parent.
- After analyzing the graph, it becomes evident that {node_type} {node} lacks a directly linked parent node.
- As depicted in the graph, {node_type} {node} exists without a parent node.
\end{lstlisting}
\caption{Prompts for the parent prediction GraphQA task.}
\label{fig:prompt-parent-prediction}
\end{figure*}

\begin{figure*}[h]
\centering
\begin{lstlisting}
Question templates:
- In this graph, what are the children of this {node_type}: {node}.
- Identify all children of {node_type} {node} in this graph.
- Find the child nodes of {node_type} {node} in the graph.
- In the graph, how many children does the {node_type} {node} have? What are they?
- How many children does {node_type} {node} have in this graph? What are they?
- Please find all children of {node_type} {node} in this graph.
- Can you find all children of {node_type} {node} in this graph?
- List all the descendant nodes of {node_type} {node} in this graph.
- What are the direct children of the {node_type} {node}?
- Can you enumerate the offspring of {node_type} {node} within this graph?
- Could you provide the list of child nodes attached to {node_type} {node}?
- Please identify the child nodes emanating from {node_type} {node}.
- Show me the child nodes of {node_type} {node}.
- What nodes are directly connected to {node_type} {node} as its children?
- I need to know all the child elements of {node_type} {node}. Can you provide that?
- Are there any nodes that directly derive from {node_type} {node} in this graph?
- Which nodes act as successors to the node tagged as {node_type} {node}?
- What are the adjacent nodes that are children of {node_type} {node}?
- Identify the nodes that are immediate successors of {node_type} {node} in this graph.
- Detail the nodes branching from {node_type} {node} in this graph structure.
- Reveal all nodes that are directly beneath {node_type} {node} in the hierarchy.

Answer templates 1 (has children):
- The given {node_type} has {child_num} children in the graph, they are: {child_nodes}
- This {node_type} has {child_num} children: {child_nodes}
- {node_type} {node} has a total of {child_num} children in this graph, which are: {child_nodes}
- There are {child_num} child nodes of {node_type} {node}, specifically: {child_nodes}
- As for the children of {node_type} {node}, you will find {child_num} direct descendants: {child_nodes}
- The count of {node_type} {node}'s children amounts to {child_num}. They include: {child_nodes}
- {node_type} {node} is parent to the following {child_num} nodes: {child_nodes}
- A list of the {child_num} children under {node_type} {node} is as follows: {child_nodes}
- Directly under {node_type} {node}, there are {child_num} children listed as: {child_nodes}
- {child_num} children spring from {node_type} {node}, which are given below: {child_nodes}

Answer tempaltes 2 (no children):
- This {node_type} does not have any child nodes in the graph.
- This {node_type} does not have any children in the graph.
- There are no children of this {node_type} in the given graph.
- The given {node_type} does not have any children in the graph.
- After examining the graph, it's determined that this {node_type} has no children.
- I've checked the {node_type} {node} and found it has no direct descendants.
- There are no child nodes attached to {node_type} {node} in this graph.
- No descendants can be traced from this {node_type}.
- The {node_type} {node} is devoid of child nodes within the current graph structure.
- It appears {node_type} {node} has no children.
\end{lstlisting}
\caption{Prompts for the child prediction GraphQA task.}
\label{fig:prompt-child-prediction}
\end{figure*}

\clearpage
\section{Statistical Tests}\label{sec:appendix-stat}
To verify the statistical significance of the main results, we conducted Chi squared tests on the four tasks with discrete performance metrics (i.e. code translation, code repair, clone detection, and defect detection) and Wilcoxon signed-rank test on the task with continuous performance metrics (i.e. code summarization). The results are presented in Table~\ref{tab:p-value}. Most of the differences are significant, except for CodeGen and Phi-1 on JavaScript summarization, and LLaMA3 on Java-to-Python translation and Python summarization.

% The complete list of p values of the statistical differences between the main results and the baselines is given in Table~\ref{tab:p-value}.

\begin{table*}[h]
    \centering
    \adjustbox{width=\textwidth,center}{
    \begin{tabular}{llccccc}
\toprule
Model & Setting & Trans (Ja2Py/Py2Ja) & Clone & Repair & Sum (Java/Py/JS) & Defect \\
\midrule
\multirow{2}{*}{CodeGen 350M} & G2C & 0.0862/0.0000 & 0.0000 & 0.0000 & 0.0114/0.0000/0.3684 & 0.0000 \\
& G2C + GQA & 0.0000/0.0001 & 0.0000 & 0.0000 & 0.0111/0.0011/0.4332 & 0.0000 \\
\midrule
\multirow{2}{*}{StarCoder 1B} & G2C & 0.0057/0.0140 & 0.0000 & 0.0000 & 0.0000/0.0000/0.0000 & 0.0000 \\
& G2C + GQA & 0.0057/0.0140 & 0.0000 & 0.0000 & 0.0000/0.0000/0.0000 & 0.0000 \\
\midrule
\multirow{2}{*}{Phi-1 1.3B} & G2C & 0.0000/0.0000 & 0.0000 & 0.0000 & 0.0090/0.0595/0.5485 & 0.0000 \\
& G2C + GQA & 0.0009/0.0000 & 0.0000 & 0.0000 & 0.0058/0.0048/0.7818 & 0.0000 \\
\midrule
\multirow{2}{*}{Qwen2.5-Coder 1.5B} & G2C & 0.0000/0.0000 & 0.0000 & 0.0000 & 0.2176/0.0001/0.0000 & 0.0000\\
& G2C + GQA & 0.0286/0.0000 & 0.0000 & 0.0000 & 0.0008/0.0003/0.0001 & 0.0000 \\
\midrule
\multirow{2}{*}{LLaMA2 7B} & G2C & 0.0037/0.0000 & 0.0000 & 0.0000 & 0.0195/0.0755/0.5454 & 0.0000 \\
& G2C + GQA & 0.0000/0.0000 & 0.0000 & 0.0000 & 0.0468/0.1406/0.0147 & 0.0000 \\
\midrule
\multirow{2}{*}{LLaMA3 8B} & G2C & 0.0003/0.0000 & 0.0000 & 0.0000 & 0.5029/0.0000/0.0014 & 0.0000 \\
& G2C + GQA & 1.0000/0.0000 & 0.0000 & 0.0000 & 0.0019/0.0147/0.0001 & 0.0000 \\
\midrule
\multirow{2}{*}{Qwen2.5-Coder 14B} & G2C & 0.0000/0.0000 & 0.0000 & 0.0000 & 0.9403/0.0213/0.0000 & 0.0000 \\
& G2C + GQA & 0.0509/0.0000 & 0.0000 & 0.0000 & 0.2508/0.6213/0.0000 & 0.0000 \\
\bottomrule
    \end{tabular}
    }
    \caption{Statistical significance (p values) of main results' differences from the baselines.}
    \label{tab:p-value}
\end{table*}

% \section{Experiments with MagNet}\label{sec:appendix-magnet}
\clearpage
\section{Example Outputs with Single-Task Finetuning}\label{sec:appendix-sft-example}
In Figure~\ref{fig:sft-hallucination}, we provide an example of model hallucination after training on only one downstream task in the GALLa framework. The model produces an answer template learned from the GraphQA task instead of responding to the actual question. This issue is mitigated in the multitask finetuning setting.

\begin{figure*}[h]
\centering
\begin{lstlisting}
** Prompt **:
# Translate the following Java program to Python

# Java
public Object splitWords(String txt) {
        if (txt.contains(" " )) {
            return Arrays.asList(txt.split(" " ));
        } else if (txt.contains("," )) {
            return Arrays.asList(txt.split("[,\s]" ));
        } else {
            int count = 0;
            for (char c : txt.toCharArray()) {
                if (Character.isLowerCase(c) && (c - 'a') % 2 == 1) {
                    count += 1;
                }
            }
            return count;
        }
    }
}

# Python
def split_words(txt):

** Expected answer **:
    if " " in txt:
        return txt.split(" ")
    elif "," in txt:
        return txt.split(",")
    else:
        count = 0
        for c in txt:
            if c.islower() and (ord(c) - ord('a')) % 2 == 1:
                count += 1
        return count

** Model Response **:
There is 1 connected component in the program's DFG.
\end{lstlisting}
\caption{An example of hallucination after single-task finetuning.}
\label{fig:sft-hallucination}
\end{figure*}

\end{document}